%% file: 01_main.tex
\documentclass[10pt,twocolumn,letterpaper]{article}

\usepackage[pagenumbers]{cvpr} %

\input{preambles/packages}

\input{preambles/macro}

\definecolor{cvprblue}{rgb}{0.21,0.49,0.74}
\usepackage[pagebackref,breaklinks,colorlinks,citecolor=cvprblue]{hyperref}

\def\paperID{****} %
\def\confName{CVPR}
\def\confYear{2024}

\title{Multi-view Image Prompted Multi-view Diffusion for Improved 3D Generation}

\author{Seungwook Kim$^{1,2}$ \hspace{0.8cm} Yichun Shi$^2$ \hspace{0.8cm} Kejie Li$^2$ \vspace{1.5mm} \hspace{0.8cm} Minsu Cho$^1$ \hspace{0.8cm} Peng Wang$^2$ \vspace{0.5cm}\\
$^1$POSTECH, South Korea \hspace{3.0cm} $^2$Bytedance, USA
}

\begin{document}

\maketitle

\input{sections/0_abstract}    
\input{sections/1_introduction}
\input{sections/2_preliminary}
\input{sections/3_method}

\input{sections/4_results}
\input{sections/5_conclusion}

\vspace{2mm}
\noindent
\small
\textbf{Acknowledgement.} 
This work was done while Seungwook Kim was an intern at Bytedance. 
Seungwook Kim was supported by the Hyundai-Motor Chung Mong-koo Foundation.

{
    \small
    \bibliographystyle{ieeenat_fullname}
    \bibliography{main}
}

\end{document}

%% file: preambles/packages.tex
\usepackage{graphicx}
\usepackage{amsmath}
\usepackage{amssymb}
\usepackage{booktabs}
\usepackage{algorithm}
\usepackage{dsfont}
\usepackage{algorithm}
\usepackage{algpseudocode}
\usepackage{algorithmicx}
\usepackage{mathtools}
\usepackage{multirow}
\usepackage{graphics}
\usepackage{wrapfig}
\usepackage{bbm}
\usepackage{xspace}
\usepackage{xcolor}
\usepackage{tikz}

%% file: preambles/macro.tex
\newcommand{\methodName}{MultiImageDream\xspace}

\usepackage{xr}
\makeatletter
\newcommand*{\addFileDependency}[1]{%
  \typeout{(#1)}
  \@addtofilelist{#1}
  \IfFileExists{#1}{}{\typeout{No file #1.}}
}
\makeatother

%% file: sections/0_abstract.tex
\begin{abstract}
Using image as prompts for 3D generation demonstrate particularly strong performances compared to using text prompts alone, for images provide a more intuitive guidance for the 3D generation process.
In this work, we delve into the potential of using multiple image prompts, instead of a single image prompt, for 3D generation.
Specifically, we build on ImageDream, a novel image-prompt multi-view diffusion model, to support multi-view images as the input prompt. 
Our method, dubbed \methodName, reveals that transitioning from a single-image prompt to multiple-image prompts enhances the performance of multi-view and 3D object generation according to various quantitative evaluation metrics and qualitative assessments. 
This advancement is achieved without the necessity of fine-tuning the pre-trained ImageDream multi-view diffusion model.
\end{abstract}

%% file: sections/1_introduction.tex
\section{Introduction}
\label{sec:intro}

Learning-based 3D object generation is an emerging field of research with rapid advancements~\cite{poole2022dreamfusion, szymanowicz2023viewset, liu2023syncdreamer, shi2023mvdream}, with potential applications to varying fields including gaming, media, or mixed reality (XR) scenarios. 
A popular trend of recent methods is to leverage the efficacy and creativity of large-scale text-to-image generative models \ie diffusion models~\cite{rombach2022high}, to generate 3D objects with high quality and fidelity. 
Among many research directions, using image inputs as prompts for 3D generation, \ie image-to-3D methods, has demonstrated strong performances albeit its simplicity~\cite{qian2023magic123, wang2023imagedream, liu2023zero123}, for image prompts serve as a means to provide rich and visual information that text alone may be insufficient or ambiguous to describe, functioning as a more effective guidance for the 3D generation process. 

The 3D generated results of image-to-3D methods exhibits realistic generation results, mostly resembling the provided image prompt at the corresponding viewpoint.
However, we observe that at other viewpoints, the generated 3D object often shows to lack details, or shows inconsistencies in texture or lighting \eg whitened texture towards the back of the object. 
Therefore, we aim to provide an insight to whether providing multi-view image prompts can alleviate the existing shortcomings of image-to-3D without compromise of overall performance. 

Noting that ImageDream~\cite{wang2023imagedream}, a recent state-of-the-art in image-to-3D generation, proposes a framework that can be seamlessly extended to support multi-view input prompts, we build on ImageDream to verify the idea of multi-image prompting for 3D generation.
Specifically, using ImageDream's best settings (\ie using pixel controller together with local controller) as the baseline, we extend the pixel and local controllers of ImageDream to support multi-image inputs.
In multi-view generation or 3D generation using SDS loss~\cite{poole2022dreamfusion}, we first use the pretrained ImageDream multi-view diffusion model to generate multiple images given a single image input, which are subsequently used as multi-image prompts for multi-view or 3D generation.
Our method, dubbed \methodName, shows to outperform ImageDream both in terms of quantitative metrics and qualitative assessment without any finetuning.

The contributions of our work are as follows:
\begin{itemize}
\item We identify that the state-of-the-art image-to-3D methods have room for improvement, especially in terms of consistent quality and texture across varying viewpoints.
\item We delve into the idea of using multiple image prompts - where multiple images can be obtained through multi-view diffusion, without having to collect ground-truth images - for improved 3D generation.
\item We show that using multiple image prompts shows quantitative and qualitative improvements even without finetuning, suggesting a promising research direction towards multi-view image prompted 3D generation.
\end{itemize}

%% file: sections/2_preliminary.tex
\section{Preliminary: ImageDream for image-to-3D}
\label{preliminary}

ImageDream~\cite{wang2023imagedream} builds on MVDream~\cite{shi2023mvdream}, a multi-view diffusion network that produces four orthogonal multi-view images given a text prompt. 
Each block of MVDream's multi-view diffusion U-Net contains a densely connected 3D attention on the four-view image latents, facilitating the learning of cross-view interactions to boost 3D consistency. 
We guide the readers to MVDream for a detailed explanation of their training scheme, which aims to improve the consistency of multi-view diffusion while maintaining the generalizability of text-to-image diffusion models.

ImageDream introduces a multi-level controller to insert the image prompt to the multi-view diffusion model at varying components of the U-Net.
The best performing version of ImageDream uses a combination of \textit{local} and \textit{pixel} controllers.
While we guide the readers to ImageDream for a detailed explanation of each controller, we briefly introduce the local and pixel controllers in the subsequent paragraphs.

\smallbreak
\noindent
\textbf{Local controller} uses the hidden feature $f_h$ from the CLIP encoder before the global pooling layer, where the feature contains detailed structural information. 
A resampling module reduces the hidden token count of $f_h$ from 257 to 16, resulting in a better-balanced local image feature $f_r$.
MLP adaptors are further incorporated to feed $f_r$ into the multi-view diffusion's cross-attention module.
However, local controller alone is insufficient to capture the finer details such as the skin texture.

\smallbreak
\noindent
\textbf{Pixel controller} aims to integrate the object appearance and texture by embedding the latent of the image prompt $x$ across all 3D dense attention layers in the multi-view diffusion model.
Specifically, the 3D dense attention of MVDream is originally applied on a stacked feature map of shape ($b,4,c,h_l,w_l$), where $b$ is the batch size, $c$ is the channel dimension, $h_l$ and $w_l$ are the spatial dimensions of the feature map, and 4 corresponds to the 4 orthogonal views.
When the pixel controller is used, the image pixel latent $x$ of the image prompt is additionally stacked, resulting in a feature map of shape ($b,4+1,c,h_l,w_l$).
This essentially enables the 3D self-attention to collectively attend to both the four orthogonal images and the input prompt image for improved pixel-level guidance.

%% file: sections/3_method.tex
\input{assets/figures/latex/main_overview}
\input{assets/figures/latex/main_qual}

\section{Method}
\label{method}

In this section, we elaborate on how we facilitate the accommodation of multi-image prompts for the pixel and local controllers of ImageDream, without fine-tuning the model.

\smallbreak
\noindent
\textbf{Multi-image local controller.}
For each of the multi-image prompts $\{I_i\}^{N}_{i=1}$, where $N$ is the total number of image prompts, their respective hidden features from the CLIP encoder before the global pooling layer, $\{f^i_h\}^{N}_{i=1}$ are all passed to the resampling module to obtain the local image features $\{f^i_r\}^{N}_{i=1}$.
Here, the total hidden token count would be $N \times 16$.
These image features are simply concatenated together before being projected to respective key and value matrices, to be subsequently fed into the cross-attention module of the multi-view diffusion.

\smallbreak
\noindent
\textbf{Multi-image pixel controller.}
Here, we aim to integrate the object appearance and texture of \textit{multiple} images.
To facilitate this, we obtain the respective pixel latents, ${x_i}_{i=1}^{N}$, and stack them altogether into a single feature map \ie the final feature shape will be $(b,N+4,c,h_l,w_l)$. 
The 3D dense attention will be carried out over this stacked feature map.

Note that the above modifications to the original ImageDream can be made seamlessly, without significant modifications to the original architecture.

%% file: assets/figures/latex/main_overview.tex
\begin{figure}[ht]
    \begin{center}
        \includegraphics[width=1.0\linewidth]{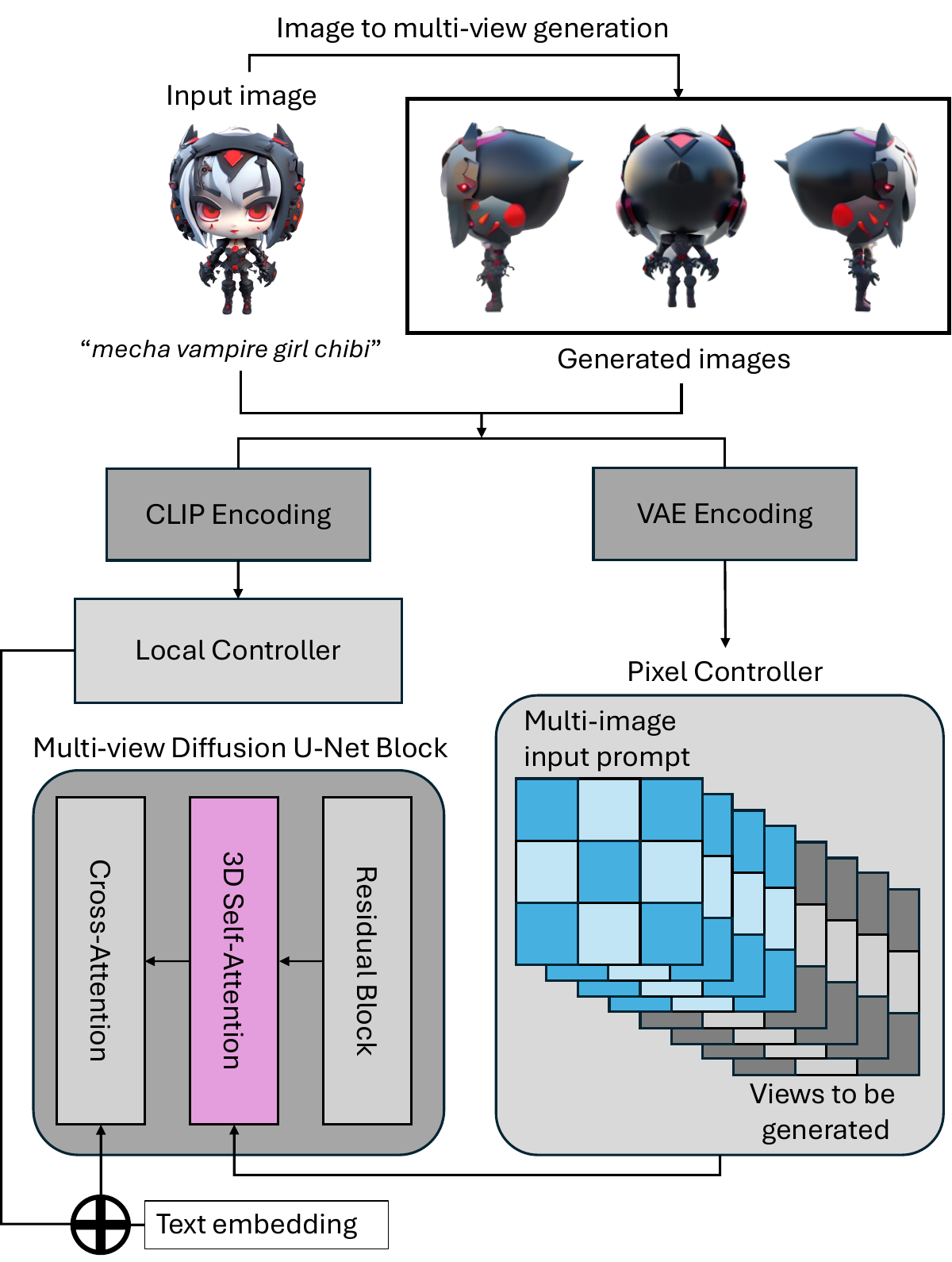}
    \end{center}
    \vspace{-5.0mm}
      \caption{\textbf{Overview of \methodName}. 
      We extend the Local and Pixel controllers proposed in ImageDream~\cite{wang2023imagedream} to support multi-image prompts for improved 3D generation.
}
\vspace{-5.0mm}
\label{fig:main_overview}
\end{figure}

%% file: assets/figures/latex/main_qual.tex
\begin{figure*}[ht]
    \begin{center}
        \includegraphics[width=0.88\linewidth]{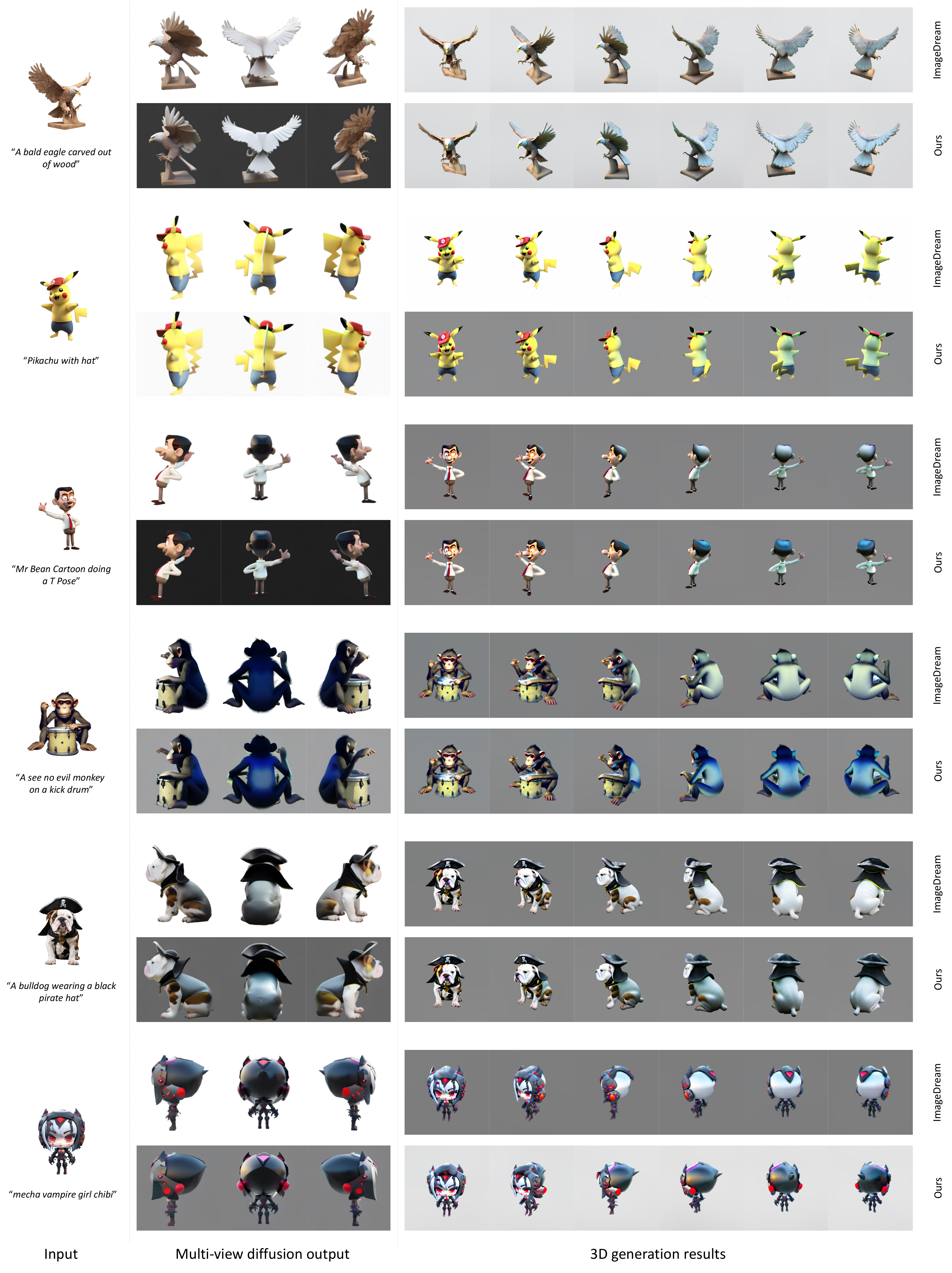}
    \end{center}
    \vspace{-5.0mm}
      \caption{\textbf{Qualitative results of \methodName in comparison to ImageDream~\cite{wang2023imagedream}}. 
      Note that the multi-view diffusion output of ImageDream are used as the additional image prompts for \methodName.
      We can observe that using multiple image prompts gives better diffusion / 3D generation outputs, alleviating issues such as excess whitening or lack of details at the back view.
      Best viewed on electronics, zoom in for better visualization.
}
\label{fig:main_qual}
\end{figure*}

%% file: sections/4_results.tex
\input{assets/tables/diffuse_inception_clip_score}
\input{assets/tables/render_inception_clip_score}
\section{Experiment}
\label{sec:experiment}

\subsection{Implementation details}
We do not conduct any additional training for \methodName; instead, we utilize the pre-trained ImageDream~\cite{wang2023imagedream} weights and extend the architecture to support multi-image prompts. 
For the training details of the image-prompted multi-view diffusion model, we direct readers to ImageDream.
For multi-view generation, our approach follows the training protocol to generate four orthogonal views corresponding to the input view.
In optimizing a NeRF model for 3D generation, we adhere to ImageDream's protocols as well.
The NeRF optimization takes about 1 hour, with only a slight increase in time when more images are fed into the pixel/local controllers, but not significantly so.

\subsection{Quantitative comparison}
For a consistent evaluation with ImageDream, we report the performance of \methodName using the Inception Score (IS)~\cite{salimans2016improved} and the CLIP scores~\cite{radford2021learning}.
We compute these scores on the same 39 prompts used in ImageDream, and thus also suffer from lack of diversity due to the small number of prompts - we therefore follow ImageDream and usethe Quality-only IS (QIS) which omits the diversity evaluation.
The results are provided in~\cref{tbl:quantitative_diffuse} and~\cref{tbl:quantitative_render}.
The baseline ImageDream is written as 1-ImageDream, as it takes just 1 image as the input prompt. 
Likewise, we name \methodName as 2-ImageDream, 3-ImageDream and 4-ImageDream, depending on the number of images they take as input.
The letters f, b, l and r refer to front, back, left and right, specifying the view direction of the input image used. 
While we evaluated on an exhaustive combination of multi-image prompts, we only report the results where using multi-image prompts outperform the baseline ImageDream in at least one of the evaluation metrics in~\cref{tbl:quantitative_diffuse}.
We then performed 3D generation using the SDS loss on only those methods, and reported the results in~\cref{tbl:quantitative_render}.
The results show that using multiple image prompts can improve the quantitative results of both multi-view generation and 3D generation, although less frequently on 3D generation.

\subsection{Qualitative comparison}
Noting that quantitative results alone are insufficient to prove the comparative efficacy of multi-view image prompting, we present qualitative comparisons in~\cref{fig:main_qual}.
It can be seen that using multi-view image prompts can alleviate several issues of single-view image prompting, such as excess whitening or lack of details at the back view.

%% file: assets/tables/diffuse_inception_clip_score.tex
\begin{table}[h]
\centering
\scalebox{0.8}{
\begin{tabular}{l |ccc}
\toprule
& \multicolumn{3}{c}{Synthesized Image} \\ 
\midrule
Model & QIS(256)$\uparrow$ & CLIP(TX)$\uparrow$ & CLIP(IM)$\uparrow$ \\ 
\midrule
SD-XL~\cite{} & 52.0$\pm$30.5 & 34.6$\pm$3.09 & 100 \\
MVDream~\cite{} & 23.05$\pm$14.4 & 31.64$\pm$2.99 & 78.41$\pm$5.32\\
\midrule
1-ImageDream~\cite{} \\
- local(f) & 22.49$\pm$9.57 & 31.32$\pm$2.86 & 82.99$\pm$6.03 \\
- pixel(f) + local(f) & 27.10$\pm$12.8 & 32.39$\pm$2.78 & 85.69$\pm$3.77 \\
\midrule
2-ImageDream \\
- pixel(f) + local(fb) & \textbf{28.95}$\pm$10.2 & \textbf{32.54}$\pm$2.65 & \textbf{87.10}$\pm$3.09 \\
- pixel(f) + local(fl) & 23.37$\pm$8.39 & \textbf{32.62}$\pm$2.76 & \textbf{86.97}$\pm$3.22 \\
- pixel(f) + local(fr) & 25.92$\pm$8.70 & \textbf{32.53}$\pm$2.64 & \textbf{87.05}$\pm$3.36 \\
- pixel(fl) + local(f) & \textbf{29.01}$\pm$14.6 & 32.06$\pm$2.84 & 85.37$\pm$3.93 \\
- pixel(fr) + local(f) & \textbf{27.94}$\pm$9.59 & 32.04$\pm$2.89 & \textbf{85.72}$\pm$3.94 \\
\midrule
3-ImageDream \\
- pixel(f) + local(fbl) & 23.48$\pm$8.81 & \textbf{32.50}$\pm$2.55 & \textbf{86.65}$\pm$3.02 \\
- pixel(f) + local(fbr) & 24.58$\pm$8.01 & \textbf{32.49}$\pm$2.63 & \textbf{86.91}$\pm$3.19 \\
- pixel(f) + local(flr) & 24.69$\pm$8.03 & \textbf{32.58}$\pm$2.56 & \textbf{86.83}$\pm$3.31 \\
\midrule
4-ImageDream \\
- pixel(f) + local(fblr) & 22.61$\pm$7.57 & \textbf{32.44}$\pm$2.61 & \textbf{86.53}$\pm$3.19\\
\bottomrule
\end{tabular}
}
\caption{
\textbf{Quantitative evaluation on diffused multi-view outputs.}
It can be seen that we can yield superior performances on certain multi-view prompt settings compared to the baseline ImageDream~\cite{wang2023imagedream} even without finetuning.
}
\label{tbl:quantitative_diffuse}
\vspace{-3mm}
\end{table}

%% file: assets/tables/render_inception_clip_score.tex
\begin{table}[h]
\centering
\scalebox{0.8}{
\begin{tabular}{l |ccc}
\toprule
& \multicolumn{3}{c}{Re-rendered} \\ 
\midrule
Model & QIS(256)$\uparrow$ & CLIP(TX)$\uparrow$ & CLIP(IM)$\uparrow$ \\ 
\midrule
MVDream~\cite{} & 29.02$\pm$10.24 & 32.69$\pm$3.39 & 79.63$\pm$4.15 \\
\midrule
1-ImageDream~\cite{} \\
- local(f) & 22.30$\pm$4.47 & 31.71$\pm$2.96 & 84.34$\pm$3.13 \\
- pixel(f) + local(f) & 25.16$\pm$6.49 & 31.59$\pm$3.23 & 84.83$\pm$4.08 \\
\midrule
2-ImageDream \\
- pixel(f) + local(fb) & 24.98$\pm$9.21 & 31.38$\pm$3.07 & 84.20$\pm$4.21 \\
- pixel(f) + local(fl) & 22.82$\pm$5.57 & 31.39$\pm$2.88 & 84.00$\pm$3.98 \\
- pixel(f) + local(fr) & 24.43$\pm$7.20 & 31.34$\pm$3.01 & 84.21$\pm$4.29 \\
- pixel(fl) + local(f) & \textbf{25.78}$\pm$5.67 & 30.74$\pm$2.64 & 83.47$\pm$4.33 \\
- pixel(fr) + local(f) & 25.14$\pm$7.82 & 31.17$\pm$2.66 & 83.92$\pm$4.41 \\
\midrule
3-ImageDream \\
- pixel(f) + local(fbl) & 23.85$\pm$7.06 & 31.23$\pm$3.07 & 83.76$\pm$4.33 \\
- pixel(f) + local(fbr) & 24.90$\pm$10.6 & 31.22$\pm$2.91 & 83.83$\pm$4.31 \\
- pixel(f) + local(flr) & 22.47$\pm$5.15 & 31.38$\pm$2.96 & 83.95$\pm$4.20 \\
\midrule
4-ImageDream \\
- pixel(f) + local(fblr) & 21.69$\pm$6.49 & 31.20$\pm$2.93 & 83.83$\pm$4.09 \\
\bottomrule
\end{tabular}
}
\caption{
\textbf{Quantitative evaluation on rendered views of the generated 3D object.}
It can be seen that we can yield competitive performances in comparison to the baseline ImageDream~\cite{wang2023imagedream} even without finetuning.
However, since quantitative metrics alone are not sufficient to evidence the superiority of one method over another in generation, we provide qualitative comparisons in~\cref{fig:main_qual}.
}
\label{tbl:quantitative_render}
\vspace{-3mm}
\end{table}

%% file: sections/5_conclusion.tex
\section{Conclusion and discussion}
\label{conclusion}
In this work, we investigated the effects of using multiple images as prompts for 3D generation, based on the observed limitations of single image-prompted 3D generation.
Our method, \methodName, builds on ImageDream, the state-of-the-art image-to-3D method, to support multi-image prompts without having to finetune the model.
\methodName shows improved 3D generation results in terms of quantitative metrics and qualitative assessment, evidencing the efficacy of using multiple images as prompts for improved 3D generation.

This work provides several promising future research directions and takeaways to motivate future studies.
While we leverage the pretrained ImageDream model, the performance of 3D generation is expected to be further improved if the model is finetuned with multi-image input prompts.
Specifically, \methodName extends ImageDream just for multi-image support, we can expect that leveraging the cross-view relations between multi-image prompts is going to provide improved generation results as well.
Also, \methodName leverages \textit{generated} images as multi-view input prompts, which are not guaranteed to be 3D consistent; delving into the effect of 3D consistency of multi-view inputs on the quality of 3D generated output is also prospected to be an important question to address.